\documentclass[review]{elsarticle}

\usepackage{hyperref}
\usepackage{tablefootnote}
\usepackage{multicol}

\usepackage{multirow}

\journal{Artificial Intelligence}

\begin{document}

\begin{frontmatter}

\title{Fine-Tuning Transformers: Vocabulary Transfer}

\author{Vladislav Mosin, Igor Samenko\fnref{foot1}}
\address{LEYA Lab, Yandex and Higher School of Economics,  3A Kantemirovskaya Street, St. Petersburg, Russia}
\fntext[foot1]{\texttt{vmos1999@gmail.com, i.samenko@gmail.com}
The publication was supported by the grant for research centers in the field of AI provided by the Analytical Center for the Government of the Russian Federation (ACRF) in accordance with the agreement on the provision of subsidies (identifier of the agreement 000000D730321P5Q0002) and the agreement with HSE University  No. 70-2021-00139. The article was prepared with the support of the Yandex DataSphere service from the Yandex Cloud platform. \\
https://cloud.yandex.com/en/services/datasphere}

\author{Borislav Kozlovskii \fnref{foot2}}
\address{Facebook UK, 1 Rathbone Square, Fitzrovia, London, UK, W1T 1HQ}
\fntext[foot2]{\texttt{kborislav@fb.com}}

\author{Alexey Tikhonov \fnref{foot3}}
\address{Independent researcher, Berlin, Germany}
\fntext[foot3]{\texttt{altsoph@gmail.com}}

\author{Ivan P. Yamshchikov \fnref{foot4}}
\address{Max Planck Institute for Mathematics in the Sciences, Leipzig, Germany \\
CEMAPRE, University of Lisbon, Lisbon, Portugal}
\fntext[foot4]{\texttt{ivan@yamshchikov.info} -- Corresponding author.}

\begin{abstract}
Transformers are responsible for the vast majority of recent advances in natural language processing. The majority of practical natural language processing applications of these models are typically enabled through transfer learning. This paper studies if corpus-specific tokenization used for fine-tuning improves the resulting performance of the model. Through a series of experiments, we demonstrate that such tokenization combined with the initialization and fine-tuning strategy for the vocabulary tokens speeds up the transfer and boosts the performance of the fine-tuned model. We call this aspect of transfer facilitation {\em vocabulary transfer}.
\end{abstract}


\end{frontmatter}

\section{Introduction}

The transformer first introduced in \cite{vaswani2017attention} is an architecture that consists of encoder and decoder stacks with stacked self-attention and point-wise, fully connected layers for both the encoder and decoder. The transformer gave rise to such models as GPT \cite{radford2018improving,radford2019language} or BERT \cite{devlin2019bert}. These architectures are shown to beat state of the art for various Natural Language Processing tasks. The performance of such models improves with the size, and training of such architectures from scratch requires a lot of computational power and huge datasets. These obstacles hinder the broader adoption of these architectures and limit most successful applications to transfer learning: a huge pretrained model is fine-tuned on a smaller dataset collected for a specific downstream task. This stimulates a growing interest to transfer learning procedures and gives rise to various approaches and practices aimed at raising the effectiveness of the transfer. For a review of transfer learning methodology for transformers, see  \cite{raffel2019exploring}. 

Typical tokenization used for transformer pretraining includes several thousand tokens. These tokens include smaller chunks of words (down to the size of a single letter) and representations with longer tokens that directly correspond to certain words. One can speculate that the model uses shorter tokens to adopt grammatical information and deal with longer, rarely observed words. In contrast, the representations with longer tokens could be useful for semantically intensive problems. These longer, semantically charged tokens may vary significantly on various downstream tasks. Therefore, adopting new downstream-specific tokenization might be beneficial for the performance of the resulting model. Indeed, various researchers have shown that corpus-specific tokenization could be beneficial for an NLP task. For example, \cite{sennrich-etal-2016-neural} show that optimal vocabulary is dependent on the frequencies of the words in the target corpus. \cite{bostrom2020byte} show that tokenization on language model pretraining has a direct impact on the resulting performance, yet do not discuss the implications of this result for transfer learning. \cite{provilkov2020bpe} introduce BPE-dropout that stochastically corrupts the segmentation procedure of Byte Pair Encoding (BPE)  \cite{sennrich-etal-2016-neural, gage1994new}, which leads to producing multiple segmentations within the same fixed BPE framework. Using BPE-dropout in the pretraining is shown to improve the downstream performance.  \cite{lakew2019controlling}, \cite{aji2020neural} and \cite{wang2021multi} discuss the tokenization in the setting of cross-language transfer.  \cite{sato2020vocabulary} demonstrate that replacing the embedding layers of the neural machine translation (NMT) model by projecting general word embeddings induced from monolingual data in a target domain onto a source-domain embedding space is beneficial for task performance. \cite{chronopoulou2020lmu} extension of the input and output embedding layer to account for the new vocabulary items improves NMT performance. Outside the NMT setting, transformer-based models are routinely fine-tuned on the same tokenization they inherit from the initial corpus. However, many NLP tasks are not cross-lingual. For example, pre-trained transformers are standardly used for text classification after fine-tuning on a task-specific corpus. Such an approach could be suboptimal since the vocabulary and frequencies of the words in a new corpus could differ significantly. This paper investigates whether new tokenization tailored for the fine-tuning corpus could improve the resulting performance of the model and speed up the transfer, and formalizes such problem as a new natural language processing task. 

If one wants a new, corpus-specific vocabulary for fine-tuning the model, one can no longer use the embedding matrix obtained in the pretraining phase. One has either learn it from scratch or come up with some fine-tuning procedures that could partially preserve the information acquired by the model in the pretraining phase. We suggest a new type of transfer learning task that we call {\em vocabulary transfer}. We define this task as finding optimal tokenization for a specific downstream task and developing such information preserving fine-tuning strategy. In this paper, we demonstrate that vocabulary transfer facilitates transfer learning in terms of downstream task quality and the speed of the transfer. To our knowledge, this is the first work that addresses the adoption of data-specific tokenization in the context of transfer-learning for transformers.

The contribution of this paper is threefold:
\begin{itemize}
\item we test several ways in which one can effectively leverage a model that was pretrained with different vocabulary tokenization;
\item we conduct a series of experiments that show that adoption of new vocabulary can indeed boost the performance of the model on the downstream tasks;
\item we thus build a case to broaden the scope of transfer learning to include a problem of fine-tuning the model on new vocabulary tokenization; we call the task addressing the effective transfer of information from an old vocabulary to a new one a {\em vocabulary transfer}.
\end{itemize}

\section{Related Work}

There are different attempts to facilitate transfer learning through some enhancement or preprocessing of the new training data. For example, \cite{arase2019transfer} proposes to inject phrasal paraphrase relations into BERT to generate suitable representations for semantic equivalence assessment instead of increasing the model's size. In this work, instead of enhancing the dataset with additional information, we try to find out if it is possible to organize transfer learning when new vocabulary tokenization is created for a fine-tuning dataset. We also want to see if there are specific ways to initialize and fine-tune new vocabulary-dependent embeddings to facilitate transfer learning. Particular works address the differences between the original vocabulary used in pretraining and the vocabulary of the dataset used for fine-tuning. For example, \cite{bojanowski2019updating} propose an algorithm that allows adapting general-purpose models to changing word distributions. \cite{sennrich-etal-2016-neural} made the neural machine translation (NMT) model capable of open-vocabulary translation by encoding rare and unknown words as sequences of subword units. \cite{kudo2018subword} address the same problem of open-vocabulary translation. The authors suggest training the model with multiple subword segmentations probabilistically sampled during training. 

It is important to note that though several subword tokenization methods are used in pretrained language models, there are only a few attempts to see how tokenization affects performance. In particular, \cite{liu2019roberta}
mention that WordPiece \cite{schuster2012japanese} has a small advantage over BPE. \cite{kudo2018subword} introduces the unigram language model tokenization
method but finds it comparable in performance to
BPE. \cite{bostrom2020byte} demonstrate that unigram-based language models such as \cite{kudo2018sentencepiece} consistently match or outperform BPE. Based on this result, in this paper, we mostly experiment with sentencepiece\footnote{https://github.com/google/sentencepiece} implementation of unigram language model tokenization developed in \cite{kudo2018sentencepiece}. To ensure that the observed effects of vocabulary transfer could be reproduced with other tokenization procedures, we also provide results with BPE and BPE-dropout tokenizations in Section \ref{sec:diftok}.

Further, we propose a method that allows us to partially inherit some knowledge from the pretrained vocabulary and pass it to new tokenization specifically tailored for downstream data. Conducting a series of experiments with several open English datasets, we demonstrate that the proposed vocabulary initialization procedure facilitates transfer learning and boosts the performance of resulting models on downstream tasks. We report our experiments with English datasets and BERT but have no reason to believe that vocabulary transfer would not be useful for other languages and models. In particular, we saw a downstream performance boost and faster transfer with GPT-2 and BERT on proprietary datasets in a more morphologically rich language. To facilitate further research on vocabulary transfer, we publish the code used for our experiments\footnote{https://github.com/LEYADEV/Vocabulary-Transfer}.

\section{Vocabulary Transfer}
\label{sec:voc}

This paper, to our knowledge, is the first to introduce the concept of vocabulary transfer. We do it as follows:

\begin{table*}[h!]
\centering
\small{\begin{tabular}{l}
 \hline
 // Randomly initialize $\widetilde{\Theta}$ , the matrix of embeddings for the new vocabulary $\widetilde{V}$\\
$\widetilde{\Theta} \leftarrow $ \verb"random_init()"\\
\\
// Copying information from old embeddings $\Theta$ for the old vocabulary $V$\\
\verb"for each new_token in" $\widetilde{V}$ \verb"do"\\
~~~\verb"if new_token exists in" $V$\\
~~~// if there is the same token in the old vocabulary, take its embedding\\
~~~~~~$\widetilde{\Theta}$[\verb"new_token"] $\leftarrow \Theta$[\verb"new_token"]\\
~~~\verb"else"\\
~~~~~~// Select some partition for the new\_token\\
~~~~~~// Calculate all tokenization of new\_token with tokens from old vocabulary $V$\\
~~~~~~\verb"partitions" $\leftarrow$ \verb"tokenize(new_token," $V$\verb")"\\
~~~~~~// Take only partitions with the smallest number of tokens\\
~~~~~~\verb"partitions" $\leftarrow$ \verb"filter(len(partition) == min(map(len,partitions)),partitions)"\\
~~~~~~// Take only partitions with the maximal length of the longest token\\
~~~~~~\verb"longest_token" $\leftarrow$ \verb"max_length(each token in partitions)"\\
~~~~~~\verb"partitions" $\leftarrow$ \verb"filter(max(map(len,partition)) == longest_token,partitions)"\\
\\
~~~~~~\verb"if len(partitions)" $>$ \verb"0"\\
~~~~~~~~~// If any multiple partitions are left after filtering, average over them\\
~~~~~~~~~\verb"for each partition in partitions do"\\
\\
~~~~~~~~~~~// Initialize the new token's embedding with average of old tokens' embeddings\\
~~~~~~~~~~~$\widetilde{\Theta}$[\verb"new_token", \verb"partition"]$\leftarrow$ \\
~~~~~~~~~~~~~~~~~~~~~~\verb"average"$( \Theta[$\verb"old_token"$]$ \verb"for each old_token in partition"$)$\\
\\ 
~~~~~~~~~~~// Average resulting embedding over all available partitions\\
~~~~~~~~~$\widetilde{\Theta}$[\verb"new_token"$]\leftarrow$ \\
~~~~~~~~~~~~~~~~~~~~~~\verb"average"$( \Theta[$\verb"new_token"$, $\verb"partition"$]$ \verb"for each partition in partitions"$)$\\
 \hline
\end{tabular}}
\caption{VIPI: Vocabulary Initialization with Partial Inheritance.}
  \label{fig:vipi}
\end{table*}


\begin{itemize}
    \item in this section, we methodologically describe vocabulary transfer as a general problem that is open to future research;
    \item we propose an example of a possible solution for the problem of vocabulary transfer;
    \item we demonstrate that such solution improves the performance of the resulting model but do not claim that such a solution is optimal;
    \item we discuss the obtained results to stimulate further interest in the problem.
\end{itemize}

Let us introduce the following notation. The vocabulary of tokens $V = \{t_k, v_k\}^{M}_{0}$  is obtained as a result of a pretraining phase. Here $t_k$ stands for some chunk of text that forms a token, and $v_k$ is an embedding that corresponds to it. The new vocabulary $\widetilde{V} =  \{\widetilde{t}_k, \widetilde{v}_k\}^{N}_{0}$ is used for the fine-tuning. 

Vocabulary transfer conceptually is a process of finding such dataset-specific tokenization  $\widetilde{V}$, its initialization, and a fine-tuning procedure for it that would result in the superior performance of a given NLP model. In the following Sections, we showcase one possible solution to the vocabulary transfer problem and run a series of ablation studies to prove that the boost in the performance is due to a specific initialization procedure and fine-tuning of the new embedding matrix.

\subsection{Example of Vocabulary Transfer: VIPI}

To transfer pretrained knowledge about old tokens to the new, corpus-specific tokens, one could have some heuristic procedure of token matching. We propose to organize such pairing in the way described in pseudo-code in Table \ref{fig:vipi}. The basic intuition behind VIPI is the attempt to preserve as much information from the old tokenization as possible in the most straightforward way. It simply searches for the strings of old tokens that form the new ones and initializes them with some average over the pretrained embeddings of old tokens.

First, if a token in the new vocabulary coincides with some token in the old one, we can assign its old embedding to it. In our experiments, from 50\% to 60\% of tokens in new, dataset-specific tokenization could be found among original tokens. At the same time, up to 30\% of new tokens could be split into a partition of several tokens from the original tokenization. For every such token in a new vocabulary, we build all possible partitions that consist of the old vocabulary tokens. Out of these partitions, we choose ones with a minimal number of tokens. If there is more than one partition with the same amount of tokens, we choose the one that includes the longest token. We iterate over these partitions if there is still more than one partition in line with these rules. We initialize the corresponding token from the new vocabulary with the old vocabulary embeddings averaged over the chosen partition for every chosen partition. Then we average these initializations across all partitions. Using this heuristic that we call VIPI or {\em Vocabulary Initialization with Partial Inheritance}, we attempt to transfer information learned on an old vocabulary to a new one. VIPI is one of the examples of vocabulary transfer that, as we show further, happens to be effective in facilitating the transfer. We hope that further research could provide better vocabulary transfer strategies.

\subsection{Data}

All experiments reported in the paper were conducted on four open datasets. We run the experiment on the BERT model that we pre-train on English Wikipedia\footnote{https://dumps.wikimedia.org/}  that has approximately sixteen gigabytes of raw text. We then use three different downstream datasets to experiment with fine-tuning: Quora Insincere Questions Detection Dataset\footnote{https://www.kaggle.com/c/quora-insincere-questions-classification/data}, Twitter sentiment analysis\footnote{http://help.sentiment140.com/} and SemEval-19 Hyperpartisan News Detection Dataset\footnote{https://pan.webis.de/semeval19/semeval19-web/}. The parameters of the datasets are shown in Table \ref{tab:dt}.

\begin{table}[h!]
\centering
\small{\begin{tabular}{lllr}
 Dataset & Size  & Documents & Unique words  \\
 \hline
 Quora & 150 Mb  & 1 306 122 & 201 122 \\
 Sentiment140  & 300 Mb  & 1 600 000 & 1 350 544 \\
 Hyper.News  & 2.2 Gb  & 735 915 & 3 688 358 \\
\end{tabular}}
\caption{Parameters of three datasets used in the experiments. Quora stands for Quora Insincere Questions Detection Dataset; Sentimen140 is a dataset for twitter sentiment analysis; Hyper.News stands for SemEval-19 Hyperpartisan News Detection Dataset.}
  \label{tab:dt}
\end{table}

All datasets were tokenized with sentencepiece\footnote{https://github.com/google/sentencepiece} implementation of Unigram Language Modelling \cite{kudo2018subword}, since according to \cite{bostrom2020byte} it is superior to BPE. We used default out-of-the-box sentencepiece tokenization for every dataset and only varied the number of tokens. We run our experiments with eight, sixteen, and thirty-two thousand tokens. The downstream datasets were randomized and split into 50\% train, 25\% dev, and 25\% test subsets, preserving the balance of labels in each subset. All results reported below are the results for the test subset unless noted otherwise.

First, BERT Masked Language Model (MLM) was trained on Wikipedia data. Then for each of the three downstream tasks, we carried out a series of experiments with various embedding initialization strategies that had the following structure:

\begin{itemize}
    \item build a new vocabulary with sentencepiece;
    \item tokenize the downstream dataset with this new vocabulary;
    \item initialize embeddings matrix of the BERT model for the new vocabulary;
    \item for every vocabulary initialization procedure run one epoch of MLM fine-tuning with new tokenization on the corresponding dataset that we transfer to;
    \item for every vocabulary initialization train the resulting BERT model classifier on a downstream dataset. 
\end{itemize}

 In Section \ref{sec:ftemb}, we show that vocabulary transfer could be directly done without the intermediary MLM step, but the resulting performance is significantly better with it, so we stick to it in other reported experiments.

\subsection{Experiments}

To see if vocabulary transfer is beneficial for transfer learning and if it can improve the resulting quality of the model, say, in terms of the downstream classifier accuracy, we compare VIPI with several different baselines. The first is the original tokenization that is fine-tuned on the downstream datasets: original tokenization, pretrained body, and embeddings. This is a standard transfer learning setup for BERT without any vocabulary transfer. The second baseline is BERT trained on the downstream dataset from scratch with the new tokenization: random body, new tokenization, and randomly initialized embeddings. The third is the model where the new embedding matrix is randomly initialized, but the body is taken from the pretrained model.  

Table \ref{tab:res} shows the resulting relative performances of all three baselines and VIPI when fine-tuning over Quora insincere questions detection, Sentiment140, and hyperpartisan news detection datasets for various numbers of tokens in the vocabulary.

\begin{table*}[]
\centering
\begin{tabular}{llllll}
\hline
\multicolumn{1}{|l|}{\multirow{3}{*}{Dataset}} & \multicolumn{1}{l|}{\multirow{3}{*}{\begin{tabular}[c]{@{}l@{}}Number of \\ tokens in \\ vocabulary \end{tabular}}} & \multicolumn{1}{l|}{\multirow{3}{*}{\begin{tabular}[c]{@{}l@{}}Original\\ Tokenization\\ pretrained\\ Body and\\ Embeddings\end{tabular}}} & \multicolumn{3}{l|}{\begin{tabular}[c]{@{}l@{}}New\\ Tokenization\end{tabular}}                                                                               \\ \cline{4-6} 
\multicolumn{1}{|l|}{}                         & \multicolumn{1}{l|}{}                                                                             & \multicolumn{1}{l|}{}                                                                                                                      & \multicolumn{1}{l|}{\begin{tabular}[c]{@{}l@{}}Random \\ Body\end{tabular}} & \multicolumn{2}{l|}{\begin{tabular}[c]{@{}l@{}}Pretrained \\ Body\end{tabular}} \\ \cline{4-6} 
\multicolumn{1}{|l|}{}                         & \multicolumn{1}{l|}{}                                                                             & \multicolumn{1}{l|}{}                                                                                                                      & \multicolumn{2}{l|}{\begin{tabular}[c]{@{}l@{}}Random \\ Embeddings\end{tabular}}                                                                   & \multicolumn{1}{l|}{VIPI}                          \\
\hline
\multirow{3}{*}{Quora, 150 Mb.}                                                        & 8 000                                                                        & 95.64                                                                                                                  & 95.82                            & 96.01         & \textbf{96.03}       \\
                                                                              & 16 000                                                                       & 95.91                                                                                                                  & 95.92                            & 96.03         & \textbf{96.11}       \\
                                                                              & 32 000                                                                       & 95.70                                                                                                                  & 95.97                            & 95.83         & \textbf{96.11}       \\ \hline
\multirow{3}{*}{\begin{tabular}[c]{@{}l@{}}Sentiment 140\\ 300 Mb.\end{tabular}}                                             & 8 000                                                                        & 85.65                                                                                                                 & 85.62                            & 85.71         & \textbf{85.73}       \\
                                                                              & 16 000                                                                       & 85.64                                                                                                                  & 85.71                            & 85.67         & \textbf{85.86}       \\
                                                                              & 32 000                                                                       & 84.53                                                                                                                  & 85.23                            & 85.78         & \textbf{85.80}       \\ \hline
\multirow{3}{*}{\begin{tabular}[c]{@{}l@{}}Hyperpartisan\\ News, 2.2 Gb.\end{tabular}} & 8 000                                                                        & 88.66                                                                                                                  & 88.72                            & 88.66        & \textbf{89.05}       \\
                                                                              & 16 000                                                                       & 86.24                                                                                                                  & 86.51                            & 88.03        & \textbf{88.58}       \\
                                                                              & 32 000                                                                       & 86.39                                                                                                                  & 86.95                            & 89.17        & \textbf{89.74}       \\ \hline
\end{tabular}
\caption{Accuracy of downstream classifiers for Quora insincere questions detection, Sentimen140 twitter sentiment classification,  and Hyperpartisan news datasets. Usage of the corpus-specific tokenization combined with the pretrained body and VIPI improves the resulting performance for all datasets across all vocabulary sizes. One could see that the effect of vocabulary transfer becomes more noticeable as the size of the downstream dataset grows.}
  \label{tab:res}
\end{table*}

\begin{figure*}
\begin{multicols}{2}
    \includegraphics[scale=0.3]{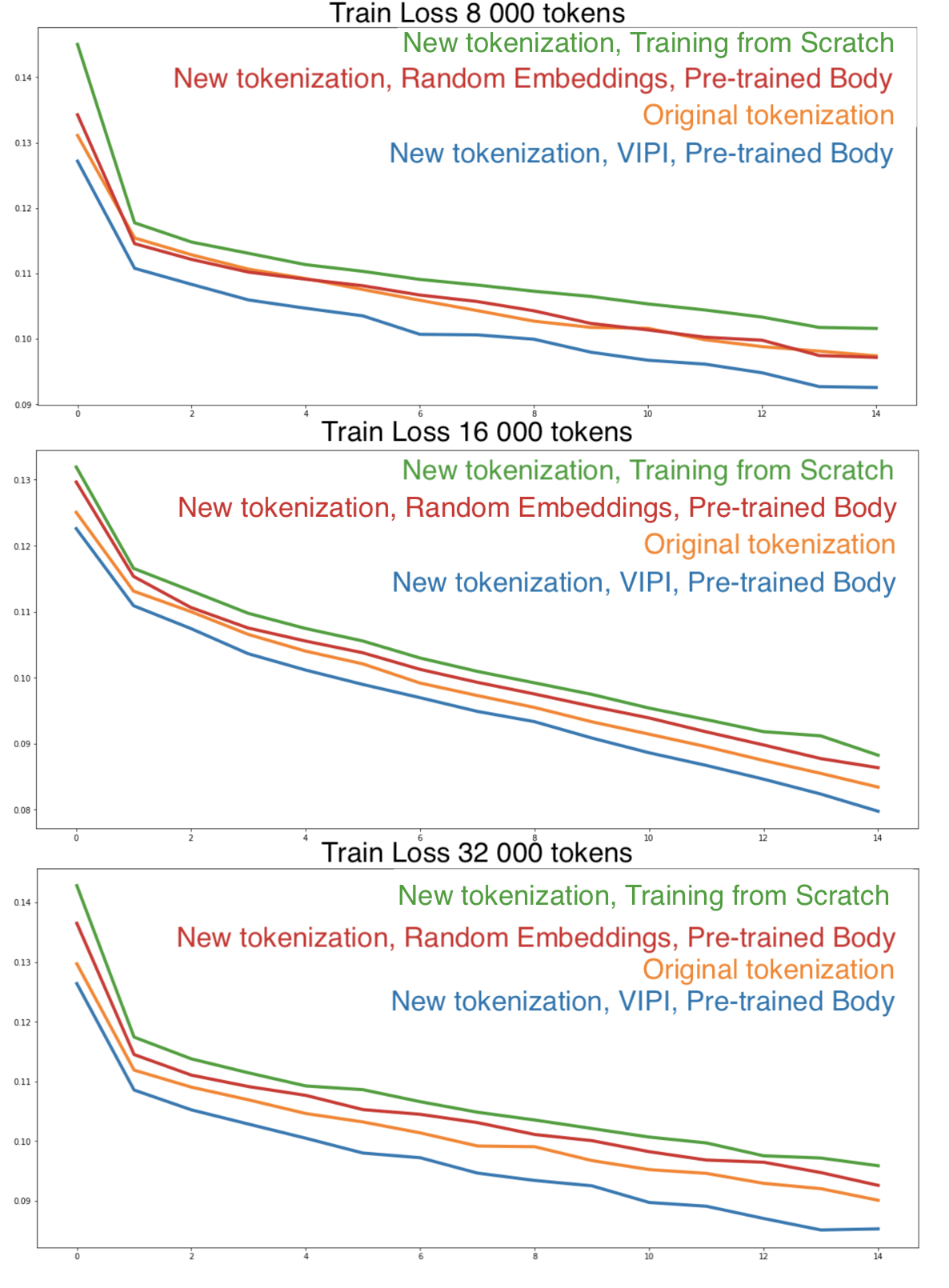}
  \caption{Quora insincere questions classifier epoch train loss. Fine-tuning the pretrained BERT with tokenization of various sizes. Corpus-specific tokenization combined with VIPI and a pretrained body speeds up learning. Models are ranked on the figure according to the resulting loss.}
  \label{fig:Q}
      \includegraphics[scale=0.295]{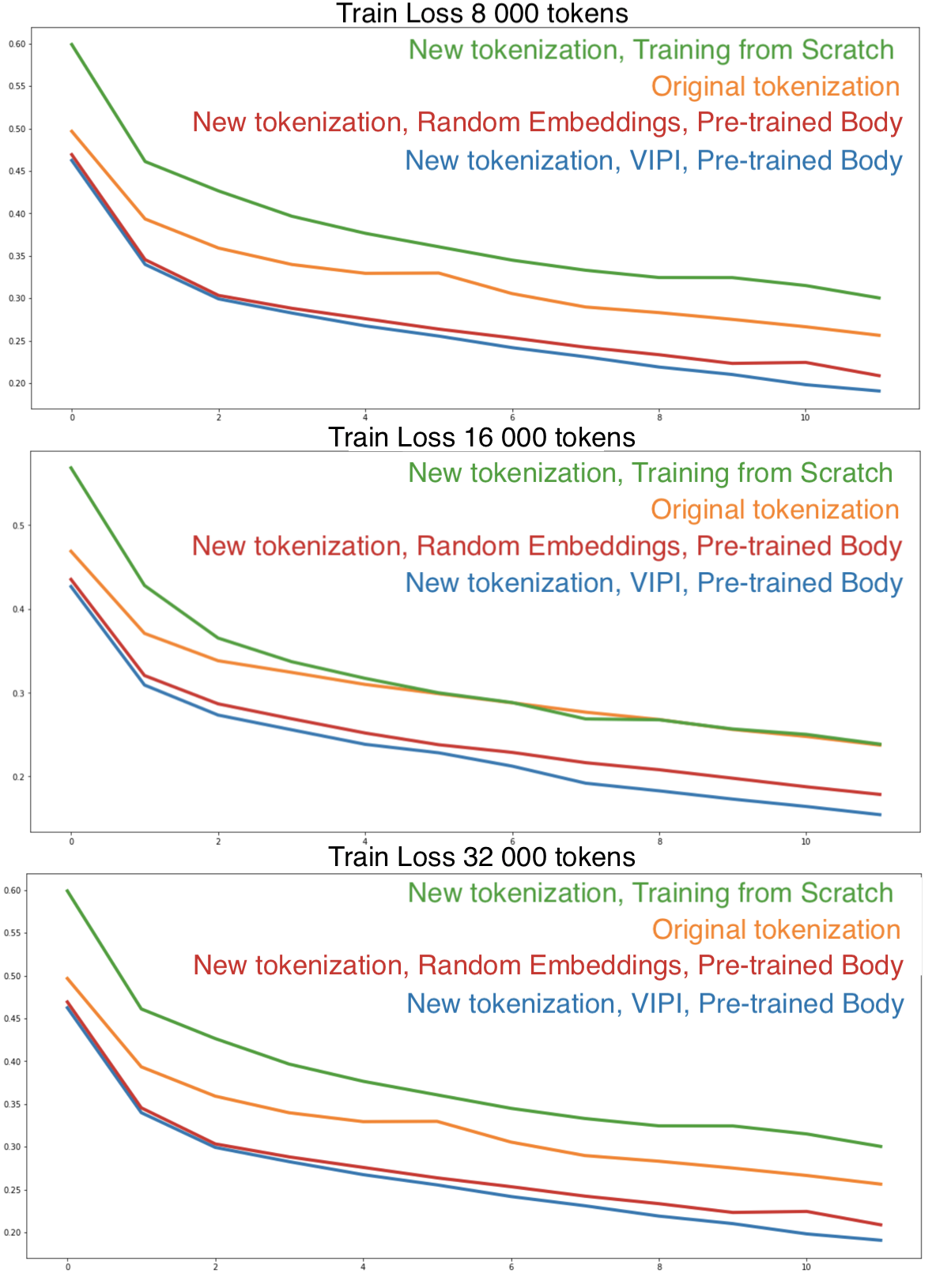}
  \caption{Hyperpartisan news classifier epoch train loss. Fine-tuning the pretrained BERT with tokenization of various sizes. The model with corpus-specific tokenizations and pretrained body learns faster than the model with original tokenization. Models are ranked on the figure according to the resulting loss.}
  \label{fig:H}
\end{multicols}
\end{figure*}

First, Table \ref{tab:res} clearly shows that new tokenization combined with VIPI outperforms all three other baselines across three different tokenization sizes. Second, Figure \ref{fig:Q} demonstrates that new tokenization combined with VIPI speeds up the transfer consistently across all three tokenization sizes. Finally, the Quora dataset is two times smaller than Sentiment140 and seven times smaller than the dataset of hyperpartisan news. Quora dataset has approximately two hundred thousand unique words; see Table \ref{tab:dt}. One could advocate that the impact of vocabulary transfer becomes more significant as the number of unique words in the dataset grows. Comparing Figure \ref{fig:H} and Figure \ref{fig:Q}, one could see that benefits of vocabulary transfer in terms of the transfer speed becomes more evident on a bigger dataset of hyperpartisan news. We address this reasoning in greater detail in Section \ref{sec:dis}.

In the next subsections, we run several ablation studies and investigate the impact of various aspects of vocabulary transfer on the resulting performance.

\section{Aspects of Vocabulary Transfer}

As we have stated in Section \ref{sec:voc} vocabulary transfer conceptually is a process of finding dataset-specific tokenization, its initialization, and fine-tuning procedure for it that would result in the superior performance of a given NLP model. We test different popular tokenizations and investigate embeddings initialization and fine-tuning procedures further in this section. 

\subsection{Embeddings Initialization}

Let us look into the initialization of the BERT's embeddings matrix in detail. First, one can claim that the tokens that coincide in both vocabularies are solely responsible for boosting the performance. To address that, we perform a series of experiments with partial initialization of the embeddings. Namely, we only initialize coinciding tokens with their pretrained values but do not complete the rest of the VIPI and initialize every other token randomly. Table \ref{tab:emb} shows that, though such initialization of matching old tokens in the vocabularies does improve the model's performance, it is inferior to VIPI. We also test two other initialization procedures. 

The first heuristic is denoted in Table \ref{tab:emb} as Matching plus Shuffle. It could be regarded as an intermediary step between matching tokens and VIPI. Here we match the tokens that coincide with the tokens present in the old vocabulary. If a new token could be split into some partition with old tokens, we randomly sample an equal amount of tokens from the old vocabulary and average across them. 

The second initialization is a version of VIPI. Instead of averaging all possible partitions, we take only one partition randomly. In Table \ref{tab:emb}, this series of experiments is denoted as VIPI over one random partition.

\begin{table*}[h]
\centering
\begin{tabular}{llllll}
\hline
\multicolumn{1}{|l|}{Dataset} & \multicolumn{1}{l|}{\begin{tabular}[c]{@{}l@{}}Number \\ of tokens\end{tabular}} & \multicolumn{1}{l|}{\begin{tabular}[c]{@{}l@{}}Random\\ Init.\end{tabular}} & \multicolumn{1}{l|}{\begin{tabular}[c]{@{}l@{}}Match\\ Old\\ Tokens\end{tabular}} & \multicolumn{1}{l|}{\begin{tabular}[c]{@{}l@{}}VIPI \\ over\\ Random\\ Partition\end{tabular}} & \multicolumn{1}{l|}{VIPI} \\ \hline            
\multirow{3}{*}{Quora}                                                        & 8 000  & 95.64 & 96.08 & \textbf{96.09} & 96.03 \\
                                                                              & 16 000 & 95.91 & 96.09 & 96.10 & \textbf{96.11} \\
                                                                              & 32 000 & 95.70 & \textbf{96.13} & 96.09 & 96.11                             \\ \hline
\multirow{3}{*}{\begin{tabular}[c]{@{}l@{}}Sentiment 140\end{tabular}} & 8 000 & 85.65 & 85.67 & \textbf{85.77} & 85.72 \\
                                                                       & 16 000 & 85.64 & \textbf{85.70} & 85.82 & 85.86 \\ 
                                                                       & 32 000 & 84.53 & \textbf{85.82} & 85.77 & 85.80   \\ \hline
                                                                              
\multirow{3}{*}{\begin{tabular}[c]{@{}l@{}}Hyperpartisan\\ News\end{tabular}} & 8 000                                                                       & 88.66                                                                                                              & 88.93                                                                                                                                                                                           & \textbf{89.05}                                                                                      & \textbf{89.05}                                      \\
                                                                              & 16 000                                                                      & 88.03                                                                                                              & 88.73                                                                                                                                                                                                          & \textbf{88.97}                                                                                       & 88.58                              \\
                                                                              & 32 000                                                                      & 89.17                                                                                                              & 89.67                                                                                                                                                                                                     & 88.99                                                                                       & \textbf{89.74}                             \\ \hline
\end{tabular}
\caption{Accuracy of downstream classifiers for Quora insincere questions detection, Sentiment140  and Hyperpartisan news datasets with corpus-specific embeddings, various embedding initialization procedures, and intermediary MLM fine-tuning. VIPI improves the resulting performance for all datasets and all vocabulary sizes.}
  \label{tab:emb}
\end{table*}

Table \ref{tab:emb} clearly illustrates several ideas. First, even such a naive idea as matching the tokens that coincide assists the transfer. Second, VIPI improves the results even further. At the same time, matching the tokens that coincide and averaging across some additional shuffled old tokens reduces the performance of the simple matching. The quality of vocabulary transfer depends on the initialization of the embeddings. In our experiments that we report further, there were approximately $30\%$ of tokens that exactly match the tokens from the initial tokenization. This fact in itself contributes to better-resulting performance. Indeed, the model performs better if it keeps the tokens it has already learned. This could be seen in the column "Match old tokens." However, these are frequent tokens that are relatively easy to learn. Thus, direct matching is sub-optimal. A more elaborate procedure for vocabulary transfer can help with the less frequent tokens. This becomes especially important on bigger datasets that include a large percentage of rarely used tokens (e.g., scientific text processing). The second case is discussed in more detail in Section \ref{sec:sc} below.

\begin{table}[h]
\centering
\begin{tabular}{llll}
\hline
\multicolumn{1}{|l|}{Number of tokens} & \multicolumn{1}{l|}{Tokenization} & \multicolumn{1}{l|}{\begin{tabular}[c]{@{}l@{}}Classification\\ Accuracy\end{tabular}} & \multicolumn{1}{l|}{\begin{tabular}[c]{@{}l@{}}Classification\\ Accuracy\\ after MLM\end{tabular}} \\ \hline
\multirow{2}{3pt}{8000}                                                      & New, VIPI                     & 86.52                                                                                               & \textbf{89.05}                                                                                                \\
                                                                            & New, Random                   & 85.22                                                                                               & 88.66                                                                                                         \\ \cline{2-4} 
                                                                            & Old Tokens                    & 84.97                                                                                               & 85.68                                                                                                         \\ \hline\hline
\multirow{2}{3pt}{16000}                                                     & New, VIPI                     & 88.45                                                                                               & \textbf{90.70}                                                                                                 \\
                                                                            & New, Random                   & 87.24                                                                                               & 90.09                                                                                                         \\ \cline{2-4} 
                                                                            & Old Tokens                    & 86.30                                                                                                & 87.62                                                                                                         \\ \hline\hline
\multirow{2}{3pt}{32000}                                                     & New, VIPI                     & 87.34                                                                                               & \textbf{89.74}                                                                                                \\
                                                                            & New, Random                   & 87.24                                                                                               & 89.17                                                                                                         \\ \cline{2-4} 
                                                                            & Old Tokens                    & 85.84                                                                                               & 86.39                                                                                                         \\ \hline
\end{tabular}
\caption{Accuracy of downstream classifiers for Hyperpartisan news dataset. Fine-tuning BERT with new, corpus-specific tokenization improves the performance with and without the intermediary MLM step. Pre-training BERT as an MLM before training of actual classifier improves the performance of all tokenization and initialization strategies. VIPI improves the resulting performance for both fine-tuning scenarios and all vocabulary sizes.}
\label{tab:Hmlmclf}
\end{table}

\subsection{Fine-tuning Embeddings} \label{sec:ftemb}

As we demonstrate above, embedding initialization plays a crucial role in vocabulary transfer, but one should also pay attention to the procedure of fine-tuning. In all experiments reported above, we initialized new embeddings with one of the heuristics, then fine-tuned BERT as an MLM, and only after this MLM step trained the downstream classifier. In this section, we look into the role of this intermediary step.

\begin{figure}[h]
\centering
     \includegraphics[scale=0.6]{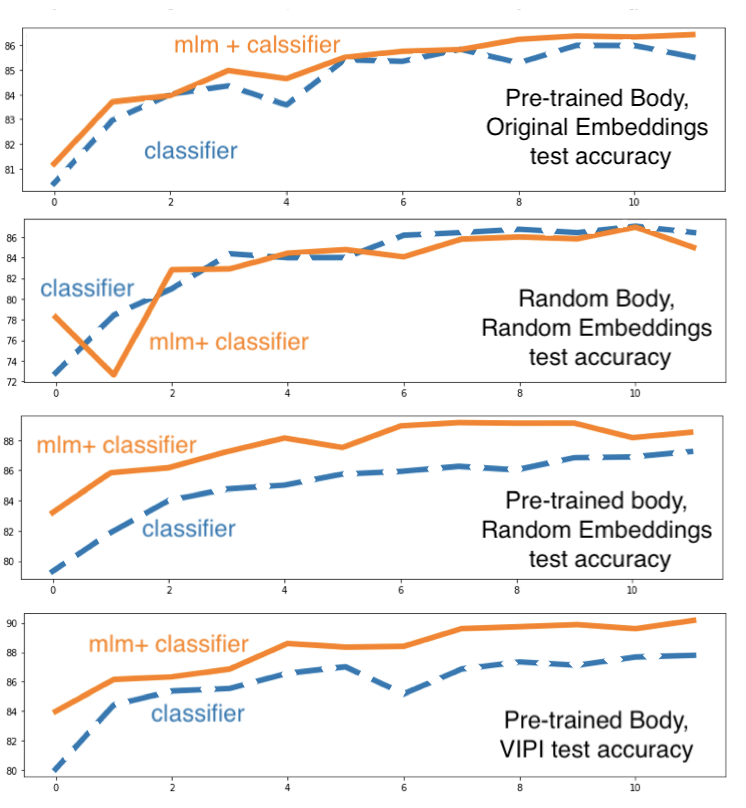}
  \caption{Fine-tuning a BERT classifier over hyperpartisan news detection data for 32 000 tokens. The first plot corresponds to the original tokenization. The second plot shows the relative performance of a classifier trained from scratch with new tokenization. The intermediary MLM step hardly plays any role when one trains a model with new tokenization from scratch. The two last plots demonstrate that pre-training an MLM model with dataset-specific tokenization boosts the performance of the resulting classifier. The classifier without MLM is represented on every plot with a dashed line.}
  \label{fig:clf}
\end{figure}

Table \ref{tab:Hmlmclf} clearly shows that adopting the new, corpus-specific tokenization benefits transfer across the board. However, the intermediary MLM step that helps to fine-tune the initialized embeddings is equally essential. Figure \ref{fig:clf} illustrates that the only case when the MLM step does not seem to be beneficial for the resulting performance is a situation when we train a classifier from scratch and do not use the pretrained body.

Summing up, we can highlight three aspects of vocabulary transfer:
\begin{itemize}
    \item new vocabulary tokenization — the size of such vocabulary and the procedure to obtain it are out of the scope of this paper but could be topics for further research;
    \item embeddings initialization procedure that could partially inherit information from the pretrained model — we propose several such initialization procedures and demonstrate that all of them outperform random initialization, and one strategy, namely, VIPI tends to be superior to others; 
    \item fine-tuning procedure for the initialized embeddings — we illustrate this aspect demonstrating that the intermediary MLM step significantly improves the adoption of new embeddings, further investigation of this aspect also looks promising.
\end{itemize}

\subsection{Experimenting with the Size of Vocabulary}

 The sizes of vocabulary in current transformer-based language models usually do not have any quantitative justification. Thus we choose three vocabulary sizes with 16 000 tokens as a "default" size and two other vocabularies that are two times bigger and two times smaller than this arbitrary default. It is reasonable to believe that the behavior observed in these vocabularies would be, to some extent, observed in the vocabularies of another size. 
 
 It makes sense to expect that the optimal vocabulary size might depend on the task in question and the data structure available for fine-tuning. While the optimal size of the vocabulary is a valid research question, it remains out of the scope of this paper. However, it might be reasonable to find out if vocabulary transfer could not only be performed over the same sized vocabulary but be used in the setup where one wants to change the size of vocabulary used for fine-tuning. In this subsection, we briefly explore this question

In order to explore possibilities of vocabulary transfer with changing vocabulary size, the applied VIPI to Quora Insincere Questions dataset using an initial vocabulary size of 16 000 tokens and fine-tuning the classifier over the vocabulary of 8 000 and 32 000 correspondingly. 

\begin{table*}[h]
\centering
\begin{tabular}{lll}
Vocabulary sizes & Change of accuracy & Change of accuracy \\
                 & for classifier + VIPI& for classifier after MLM + VIPI\\
\hline
$16 000 \rightarrow 8 000$ & +1.7\%& +1.7\%\\
$16 000 \rightarrow 32 000$ & +1.3\% &+2.3\%\\
\hline
\end{tabular}
\caption{Relative change in accuracy of downstream classifiers for Quora  Insincere Questions dataset in comparison with standard fine-tuning baseline with 16 000 tokens. Fine-tuning BERT with new, corpus-specific tokenization improves the performance with and without the intermediary MLM step, even if the size of the vocabulary is changed. The effect is more pronounced when vocabulary transfer is performed from a smaller vocabulary to a bigger one. Pre-training BERT as an MLM before training of actual classifier improves the performance on larger vocabulary yet has no benefits when we transfer to a vocabulary of a smaller size.}
\label{tab:16832}
\end{table*}

Looking at \ref{tab:16832}, one could see that vocabulary transfer could improve the resulting performance even when the size of the target vocabulary differs from the size of the initial one when compared with standard fine-tuning of the model with the fixed vocabulary size. It also stands to reason that injecting a larger vocabulary into the vocabulary of a smaller size is easier, and an additional MLM step hardly improves the resulting performance. However, an intermediary MLM step significantly improves the downstream performance on the dataset with a larger vocabulary.

\begin{table*}[h]
\centering
\begin{tabular}{lllllll}
\hline
\multicolumn{1}{|l|}{Dataset} & \multicolumn{1}{l|}{\begin{tabular}[c]{@{}l@{}}Number \\ of tokens\end{tabular}} & \multicolumn{1}{l|}{\begin{tabular}[c]{@{}l@{}}Random\\ Init.\end{tabular}} & \multicolumn{1}{l|}{\begin{tabular}[c]{@{}l@{}}Match\\ Old\\ Tokens\end{tabular}} & \multicolumn{1}{l|}{\begin{tabular}[c]{@{}l@{}}VIPI \\ over\\ Random\\ Partition\end{tabular}} & \multicolumn{1}{l|}{VIPI} \\ \hline            
\multirow{3}{*}{Quora}                                                        & 8 000  & 96.02 & 96.05 & 96.02 & \textbf{96.06} \\ 
                                                                              & 16 000 & 95.86 & 96.08 & \textbf{96.10} & 96.06 \\
                                                                              & 32 000 & 95.68 & 96.03 & 96.08 & \textbf{96.10}                             \\ \hline
\multirow{3}{*}{\begin{tabular}[c]{@{}l@{}}Sentiment 140\end{tabular}} & 8 000 & 85.56 & 85.56 & 85.65 & \textbf{85.71} \\
                                                                       & 16 000 & 85.56 & \textbf{85.77} & 85.73 & 85.67 \\
                                                                      & 32 000 & 85.72 & 85.67 & 85.70 & \textbf{85.91}   \\ \hline
\end{tabular}
\caption{Accuracy of downstream classifiers for Quora insincere questions detection, Sentiment140  and Hyperpartisan news datasets with corpus-specific embeddings, various embedding initialization procedures, and intermediary MLM fine-tuning with BPE tokenization.}
  \label{tab:embBPE}
\end{table*}

\begin{table*}[h]
\centering
\begin{tabular}{lllllll}
\hline
\multicolumn{1}{|l|}{Dataset} & \multicolumn{1}{l|}{\begin{tabular}[c]{@{}l@{}}Number \\ of tokens\end{tabular}} & \multicolumn{1}{l|}{\begin{tabular}[c]{@{}l@{}}Random\\ Init.\end{tabular}} & \multicolumn{1}{l|}{\begin{tabular}[c]{@{}l@{}}Match\\ Old\\ Tokens\end{tabular}} & \multicolumn{1}{l|}{\begin{tabular}[c]{@{}l@{}}VIPI \\ over\\ Random\\ Partition\end{tabular}} & \multicolumn{1}{l|}{VIPI} \\ \hline            
\multirow{3}{*}{Quora}                                                        & 8 000  & 95.95 & 95.95 & 95.94 & \textbf{95.96} \\
                                                                              & 16 000 & 95.95 & 95.94 & \textbf{96.03} & 96.00 \\
                                                                              & 32 000 & 96.01 & 95.94 & \textbf{96.05} & 95.98                     \\ \hline
\multirow{3}{*}{\begin{tabular}[c]{@{}l@{}}Sentiment 140\end{tabular}} & 8 000 & 85.26 & 85.31 & \textbf{85.42} & 85.40 \\
                                                                       & 16 000 & 85.52 & 85.48 & 85.45 & \textbf{85.58} \\
                                                                       & 32 000 & 85.41 & 85.56 & \textbf{85.69} & 85.65   \\ \hline
\end{tabular}
\caption{Accuracy of downstream classifiers for Quora insincere questions detection, Sentiment140  and Hyperpartisan news datasets with corpus-specific embeddings, various embedding initialization procedures, and intermediary MLM fine-tuning with BPE-Dropout tokenization.}
  \label{tab:embBPEDropout}
\end{table*}

\subsection{Experiments with different tokenizations}
\label{sec:diftok}
Experiments in the main part of this paper were processed with the most common tokenization technique - Unigram Language Modelling \cite{kudo2018subword}, with results presented in Table \ref{tab:emb}. To prove that vocabulary transfer is not an attribute of a specific tokenization we provide results on Quora and Sentiment 140 with two more popular tokenizations: Byte-Pair Encoding (BPE) \cite{sennrich-etal-2016-neural} and BPE-Dropout \cite{provilkov2020bpe}, see Table \ref{tab:embBPE} and Table \ref{tab:embBPEDropout} respectively. Experiments show that performance is similar to the results in the previous sections. Thus vocabulary transfer positive effect may be detected for different tokenizations.

\subsection{Vocabulary Transfer for Scientific Text Processing} \label{sec:sc}

As we have already mentioned, vocabulary transfer seems to be especially relevant when the dataset used for fine-tuning has rarely used words. In this case, better transfer of the existing knowledge could be especially relevant. For example, "entomophobia" is a specific phobia characterized by an excessive or unrealistic fear of one or more classes of insects. Though it is reasonably rare in generic texts, its' proper disambiguation might be relevant for medical purposes. Let us check if vocabulary transfer improves the performance of the model on such datasets that significantly differ from generic natural language. For example, we can use an annotated dataset of medical texts presented in \cite{Wen2020MeDALMA}. Table \ref{tab:med} summarizes the performance of BERT with a vocabulary of 16 000 tokens with and without VIPI on the task of medical diagnosis prediction. Indeed, the effect of vocabulary transfer on medical text seems even more pronounced. Thus, we expect that vocabulary transfer would be especially relevant for scientific text processing.

\begin{table*}[h]
\centering
\begin{tabular}{lrr}
 Fine-tuning & Classifier Accuracy & MLM + Classifier Accuracy \\
\hline
 Standard &77.4 & 80.4\\
 VIPI & \textbf{81.9} & \textbf{83.1}\\
\hline
\end{tabular}
\caption{Accuracy of downstream classifiers for MeDAL. Fine-tuning BERT with new, corpus-specific tokenization improves the performance with and without the intermediary MLM step. Pre-training BERT as an MLM before training of actual classifier improves the performance further.}
\label{tab:med}
\end{table*}

\section{Discussion}
\label{sec:dis}

 Though fine-tuning BERT over new tokenization seems beneficial across various vocabulary sizes and speeds up the transfer for both downstream datasets, one has to address the discrepancy between the benefits of vocabulary transfer for hyperpartisan news detection or Sentiment140, and Quora insincere questions datasets. Indeed, Table \ref{tab:res} shows that vocabulary transfer is much more useful for hyperpartisan news than for Quora. Let us briefly discuss this difference since it provides an illustrative example of applicability for vocabulary transfer.
 
\begin{figure}[h]
 \centering
     \includegraphics[scale=0.36]{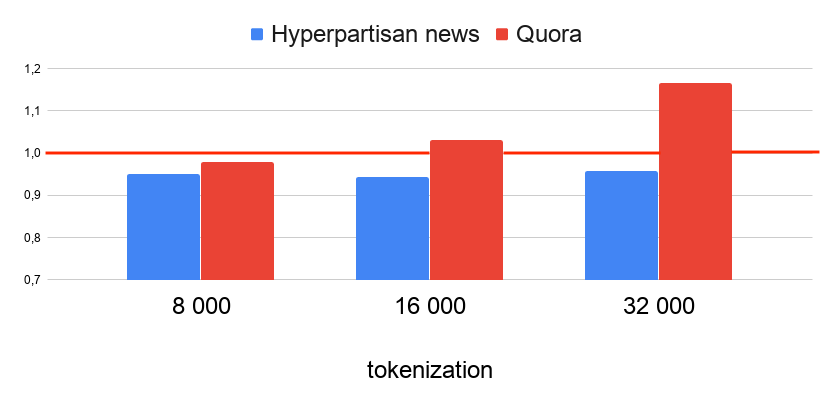}
  \caption{Relative information compression with data-specific tokenization for Quora insincere questions and hyperpartisan news datasets. The ratio between information estimators for dataset-specific tokenization and the same information estimator for original tokenization. Dataset-specific tokenization compresses hyperpartisan news datasets across various vocabulary sizes. Dataset-specific tokenization for Quora does not compress data for larger vocabularies and generally is weaker than for other datasets.}
  \label{fig:comp}
\end{figure}

 Let us return to Table \ref{tab:dt} and look at it carefully. There are several differences between the three downstream datasets. First of all, dataset sizes differ significantly: the Quora dataset is seven times smaller than hyperpartisan news and two times smaller than Sentiment140. At the same time, it has twice as many documents as hyperpartisan news. This stands to reason: questions tend to be shorter than news articles. Finally, hyperpartisan news contains fifteen times more unique words. These differences might be important for dataset-specific tokenization. One would intuitively expect that such tokenization would have a higher impact on the hyperpartisan news dataset. Figure \ref{fig:comp} proves this intuition.
 
To estimate the impact of dataset-specific tokenization on the data representation, we propose a very naive heuristic. Take a part of the downstream dataset, tokenize it, and calculate self-information simply as $-log(1/f_k)$, where $f$ stands for the frequency of the token. Let us sum this self-information across the test set. With this estimator, one could see if the dataset-specific tokenization helps to compress data: take a ratio of the self-information estimator for the dataset-specific tokenization $I_{\widetilde{V}}$ to the self-information estimator for the original tokenization $I_{V}$. One could say that if $\frac{I_{\widetilde{V}}}{I_{V}} < 1$ the dataset-specific tokenization facilitates information processing. Comparing the resulting performance shown in Table \ref{tab:res} and information compression in Figure \ref{fig:comp} one could see that this simple heuristic allows estimating the applicability of vocabulary transfer. If corpus-specific tokenization compresses the downstream dataset more effectively than original tokenization, one could expect that vocabulary transfer would have an impact on the downstream performance. 


We hope that further research into vocabulary transfer to facilitate transfer learning could establish its other properties. In particular, it is interesting to find out if the behavior shown in this paper is a consequence of specific aspects related to transformer architecture or if a general property of NLP problems leads to such behavior. Another line that we encourage is further research into how various tokenization procedures could affect downstream performance.

\section{Conclusion}

 This paper studies the effect of dataset-specific tokenization on the fine-tuning of a transformer-based architecture. We carry out experiments that demonstrate that a dataset-specific vocabulary paired with procedures for the initialization and fine-tuning of the embeddings facilitates transfer learning. We call this phenomenon {\em vocabulary transfer}. 
 
 We discuss three aspects of vocabulary transfer: tokenization, initialization, and fine-tuning. We demonstrate that dataset-specific tokenization is helpful for different datasets and fine-tuning procedures. We propose several embedding initialization procedures and show that one of them, Vocabulary Initialization with Partial Inheritance (VIPI), is superior to the others. Through a series of experiments, we demonstrate that a fine-tuning scenario for the model with new tokenization has a comparable influence on the performance as the initialization strategy. To our knowledge, these are the first results of such type. Finally, we outline the further direction of research around vocabulary transfer.

\bibliography{acl2020}

\end{document}